# Title

Is sub-metre resolution necessary for cocoa mapping? A landscape-stratified evaluation of very high resolution imagery, decametric Earth Observation inputs, and operational products in Côte d'Ivoire

# Authors

Kasimir Orlowski[1,2], Filip Sabo[3,*], Michele Meroni[4], Astrid Verhegghen[5,6], Mariana Belgiu[1], Felix Rembold[3]

[1]Faculty of Geo-Information Science and Earth Observation (ITC), University of Twente, Hallenweg 8, 7522 NH Enschede, The Netherlands.
[2]FINCONS S.P.A., Via Torri Bianche 10, Palazzo Betulla, 20871 Vimercate, Italy.
[3]European Commission, Joint Research Centre (JRC), Via Enrico Fermi, 2749, 21027 Ispra, Italy.
[4]Seidor Consulting, Carrer de Veneçuela 74, 08019 Barcelona, Spain.
[5]ARHS Developments Italia S.R.L., Via Fratelli Gabba 1/A, 20121 Milan, Italy.
[6]Flemish Institute for Technological Research (VITO), Boeretang 200, 2400 Mol, Belgium.

*Corresponding author: filip.szabo@ec.europa.eu

# Abstract

Accurate cocoa mapping is increasingly important for deforestation monitoring, supply-chain transparency, and regulatory applications. Spatial aggregation in conventional medium-resolution Earth observation (EO) imagery may limit cocoa detection in heterogeneous smallholder landscapes. In Côte d'Ivoire, we therefore evaluated how mapping performance varies across landscape conditions, whether very high resolution (VHR) imagery provides a meaningful advantage, and whether foundation-model embeddings improve decametric cocoa mapping.

We developed models using 0.5 m Pléiades VHR imagery, a 10 m Sentinel-2 annual composite, and embeddings from TESSERA and AlphaEarth Foundations (AEF), and additionally assessed four publicly available cocoa mapping products. Performance was evaluated through a landscape-stratified accuracy assessment using 2,821 independently interpreted reference points distributed across gradients of tree cover density and landscape fragmentation.

The VHR model achieved the highest performance (F1 = 0.92) and maintained F1-scores above 0.90 across all strata. Among the decametric inputs, TESSERA performed best (F1 = 0.86), followed by AEF (F1 = 0.82) and Sentinel-2 (F1 = 0.76). Of the existing cocoa products, the Kalischek product performed best (F1 = 0.83), comparable to the internally trained AEF model.

Performance differences between VHR and decametric approaches increased with fragmentation and under low and high tree cover density conditions. Targeted VHR acquisition may therefore be particularly beneficial in complex cocoa landscapes, while foundation-model embeddings offer a scalable alternative for large-area mapping.

# 1. Introduction

Cocoa cultivation supports the livelihoods of millions of smallholder farmers across tropical regions while simultaneously representing one of the most important drivers of tropical deforestation in West Africa (Squicciarini and Swinnen, 2016; Kalischek et al., 2023; Masolele et al., 2024; Renier et al., 2025; Singh and Persson, 2026). Historically, cocoa expansion has followed forest frontier dynamics, where newly cleared land initially provides high yields that decline over time as soils degrade and plantations age (Clough et al., 2009; Ruf and Siswoputranto, 1995), triggering further expansion and deforestation. Côte d'Ivoire, the world's largest cocoa producer, exemplifies this pattern: the country contained an estimated 44,500 $km^2$ of cocoa cultivation in 2021 and has experienced extensive cocoa-associated forest loss over recent decades (Kalischek et al., 2023; Masolele et al., 2024; Renier et al., 2025).

The European Union Regulation on Deforestation-Free Products (EUDR, Regulation (EU) 2023/1115) requires that cattle, cocoa, coffee, palm oil, soya, wood, and rubber placed on the EU market are not associated with deforestation after 31 December 2020 (European Union, 2023). Compliance obligations require operators to provide georeferenced data information on commodity sourcing locations and to demonstrate that production did not occur on deforested land after this date, thereby substantially increasing the operational relevance of Earth Observation (EO)-based cocoa mapping for deforestation monitoring, supply-chain due diligence, and regulatory risk assessment (Berger et al., 2025; European Union, 2023).

Recent years have seen rapid progress in EO-based cocoa mapping, particularly in Côte d'Ivoire and Ghana. Published approaches employ a wide range of EO inputs, modelling strategies, and training data sources ranging from Random Forest classifiers applied to Sentinel-1 (S1) and Sentinel-2 (S2) composites (Abu et al., 2021) to deep learning approaches based on Sentinel time series (Kalischek et al., 2023) and multi-sensor EO data (Therias et al., 2025), often trained using proprietary reference datasets. Several cocoa mapping products at national scale are now publicly accessible, including the BNETD Côte d'Ivoire land cover map (BNETD-CIGN, 2024), the Kalischek et al. (2023) cocoa probability layer, and the Forest Data Partnership (FDaP) cocoa community model layers (Forest Data Partnership, 2025, 2026).

Despite this progress, cocoa mapping remains challenging in heterogeneous tropical landscapes. Cocoa cultivation systems range from full-sun monocultures to structurally complex agroforestry systems containing varying densities of overstory shade trees (Becker et al., 2025; Clough et al., 2009; Niether et al., 2020; Vaast and Somarriba, 2014). In many producing regions, cocoa is cultivated within fragmented smallholder landscapes where cocoa plots are interspersed with other crops, fallows, secondary vegetation, and forest remnants (Schneider et al., 2023; Schroth et al., 2016). Such fine-scale heterogeneity can produce mixed pixels and reduce class separability at medium spatial resolutions (e.g., 10 m S2; Lu and Weng, 2007). At such resolutions, cocoa-specific studies have reported persistent confusion between shaded cocoa agroforestry systems and forest, including in approaches using SAR-based texture information and deep learning (Numbisi et al., 2019; Therias et al., 2025).

Cocoa mapping errors are unlikely to be uniform across landscapes. In young or sparsely planted cocoa systems, cocoa canopy cover may represent only a small proportion of the observed signal within a 10 m pixel, causing spectral responses to be dominated by soil, herbaceous vegetation, or other background components and thereby increasing omission errors. In contrast, dense agroforestry systems can be confused with forest because overstory canopy cover may partially or fully obscure the cocoa signal (Therias et al., 2025), potentially increasing both omission and commission errors. These contrasting error sources suggest that the performance of any cocoa mapping approach is likely to depend systematically on landscape structure, motivating an evaluation that accounts explicitly for variation in canopy conditions and landscape fragmentation. Examples of sparse, full-sun monoculture, and agroforestry cocoa systems and their contrasting representation in very high resolution (VHR; 0.5 m) and 10 m S2 imagery are shown in Figure 1.

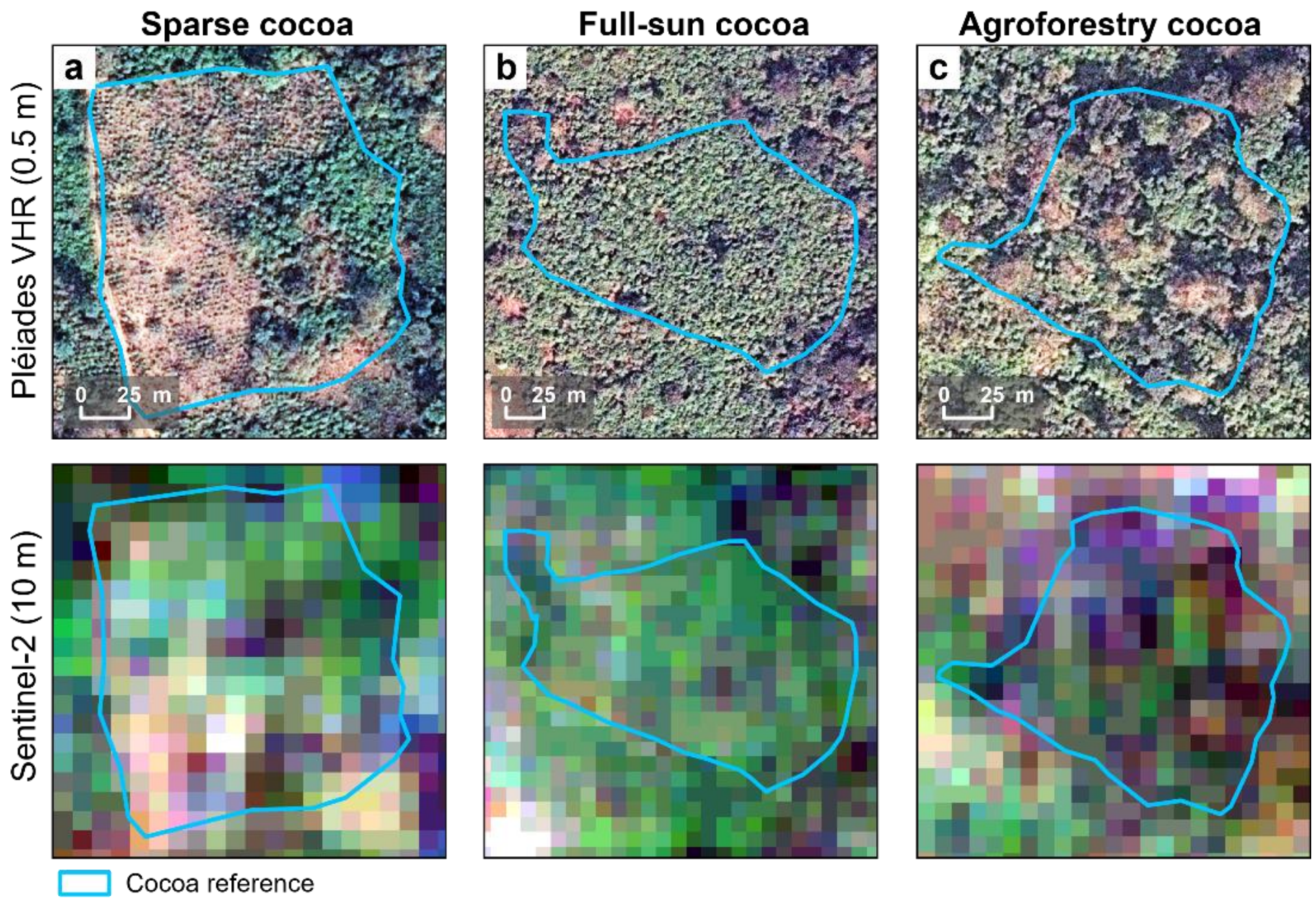

**Figure 1.** Contrasting cocoa production systems shown in 0.5 m Pléiades very high resolution (VHR) imagery and 10 m Sentinel-2 (S2) imagery: (a) sparse cocoa, (b) full-sun cocoa monoculture, and (c) agroforestry cocoa. Cyan outlines indicate the cocoa reference polygons from the COLD-CI dataset (Orlowski et al., 2026). Corresponding upper and lower panels show the same spatial extent. Both VHR and S2 imagery are displayed as true-colour composites and were individually contrast-stretched for visualisation. The examples illustrate how fine-scale canopy structure, gaps, and planted-area boundaries visible in VHR imagery are aggregated within S2 pixels. VHR image data: © CNES (2020), distribution Airbus Defence and Space.

Sub-metre VHR imagery offers spatial detail far beyond medium-resolution data, which may reduce mixed-pixel effects and improve the representation of fine-scale canopy structure, crown arrangement, and field boundaries in fragmented smallholder landscapes (Lu and Weng, 2007; Möhring et al., 2025; Zhu et al., 2026). However, VHR acquisitions involve important trade-offs relative to medium-resolution observations: coverage is typically constrained to tasked or pre-existing acquisitions, and the cost of systematic repeated acquisitions limits the temporal depth available in practice. Cloud contamination, while affecting all optical sensors, has a proportionally greater impact on VHR temporal sampling because the limited number of available acquisitions leaves fewer cloud-free observations to draw from. VHR-based approaches therefore typically rely on a small number of acquisitions rather than the dense multi-temporal stacks available from archives such as S2, and potential advantages are consequently likely to arise from improved spatial representation rather than enhanced spectral or temporal discrimination. Even so, increased spatial resolution does not resolve all cocoa mapping ambiguities. In sparse or young plantations, the spatial arrangement and contextual information captured by VHR imagery may partially compensate for a weak cocoa-specific spectral signal. In dense agroforestry systems, however, optical imagery, even at sub-metre resolution, primarily captures the visible upper canopy, so overstory trees can partly or fully obscure the cocoa layer beneath. Because VHR and medium-resolution inputs differ not only in spatial resolution but also in the spectral and temporal information they provide, observed performance differences between them cannot be attributed to spatial resolution alone.

These limitations have motivated growing interest in an emerging class of EO representations: embeddings generated by self-supervised geospatial foundation models, which may offer a practical alternative to conventional EO inputs, including medium-resolution composites and VHR imagery. Models such as TESSERA (Feng et al., 2025) and AlphaEarth Foundations (AEF; Brown et al., 2025) generate compact embeddings from multi-sensor EO archives that encode spectral patterns and temporal dynamics while reducing the need for task-specific preprocessing and feature engineering. Because these embeddings are derived from dense, wall-to-wall medium-resolution archives, they retain the spatial coverage and historical availability that VHR imagery lacks. Recent studies suggest that such embeddings can outperform conventional Sentinel-based approaches across several EO tasks while requiring comparatively limited labelled data (Ball et al., 2026; Feng et al., 2026; Lisaius et al., 2026a). However, their performance relative to VHR imagery has not yet been systematically evaluated for cocoa mapping.

Comparing different cocoa mapping approaches rigorously remains challenging. Existing cocoa mapping studies often differ simultaneously in training data, model architecture, temporal coverage, preprocessing procedures, and evaluation protocols, making it difficult to isolate the factors driving performance differences. Furthermore, most studies report global accuracy metrics that may obscure systematic performance variation across different landscape conditions. This is a particular concern given the possibly contrasting error profiles associated with varying canopy conditions described above. As a result, it remains unclear under which landscape conditions increased spatial resolution materially improves cocoa mapping performance, whether foundation-model embeddings can compensate for lower spatial resolution, and where the additional costs of VHR imagery are operationally justified.

This study addresses these gaps through a landscape-stratified comparison of cocoa mapping approaches in Côte d'Ivoire. We compare internally trained models based on 0.5 m Pléiades VHR imagery, a conventional S2 annual composite, and two foundation-model embeddings, and evaluate four existing cocoa mapping products for operational context. Specifically, we ask: (i) how landscape structure influences the performance of different EO input types; (ii) whether VHR imagery provides meaningful advantages over decametric EO inputs and, if so, under which landscape conditions; and (iii) whether foundation-model embeddings improve upon conventional S2 inputs and how their performance compares with that of VHR imagery.

# 2. Materials and Methods

## 2.1 Study area

The study was conducted in Côte d'Ivoire, where cocoa cultivation is concentrated primarily within the humid and sub-humid regions of the southern and central parts of the country (Kalischek et al., 2023). Production systems range from intensively managed full-sun monoculture plantations to structurally complex agroforestry systems with varying shade tree densities (Becker et al., 2025; Clough et al., 2009; Niether et al., 2020; Vaast and Somarriba, 2014).

The study area encompasses diverse cocoa-producing landscapes characterised by substantial variation in canopy cover, landscape fragmentation, and surrounding land cover composition (Figure 2). Cocoa-producing areas occur within heterogeneous mosaics containing forest remnants, secondary vegetation, fallows, agricultural land, oil palm and rubber plantations, settlements, and other land cover types (BNETD-CIGN, 2024). This heterogeneity creates challenging conditions for EO-based cocoa mapping because cocoa signals may vary substantially across landscapes depending on canopy closure, shade-tree density, plantation age, and adjacency to other vegetation types.

Côte d'Ivoire further provides a suitable environment for comparative cocoa mapping research because multiple EO-based cocoa mapping products are available at national scale, including cocoa probability layers and national land cover products (BNETD-CIGN, 2024; Forest Data Partnership, 2026; Kalischek et al., 2023). Such dense overlap between publicly available and operational cocoa products remains limited in other major cocoa-producing countries, enabling a broader comparative evaluation framework in the Ivorian context.

The spatial extent of the study area (Figure 2) was constrained by the availability of reference data and overlapping VHR imagery. In particular, the analysis focused on areas covered by the COLD-CI dataset (Orlowski et al., 2026), which currently provides the only large-scale cocoa and non-cocoa polygon reference dataset for Côte d'Ivoire specifically designed for EO applications. The dataset covers multiple cocoa-producing regions and landscape types, enabling the comparison of different cocoa mapping approaches within a consistent experimental framework.

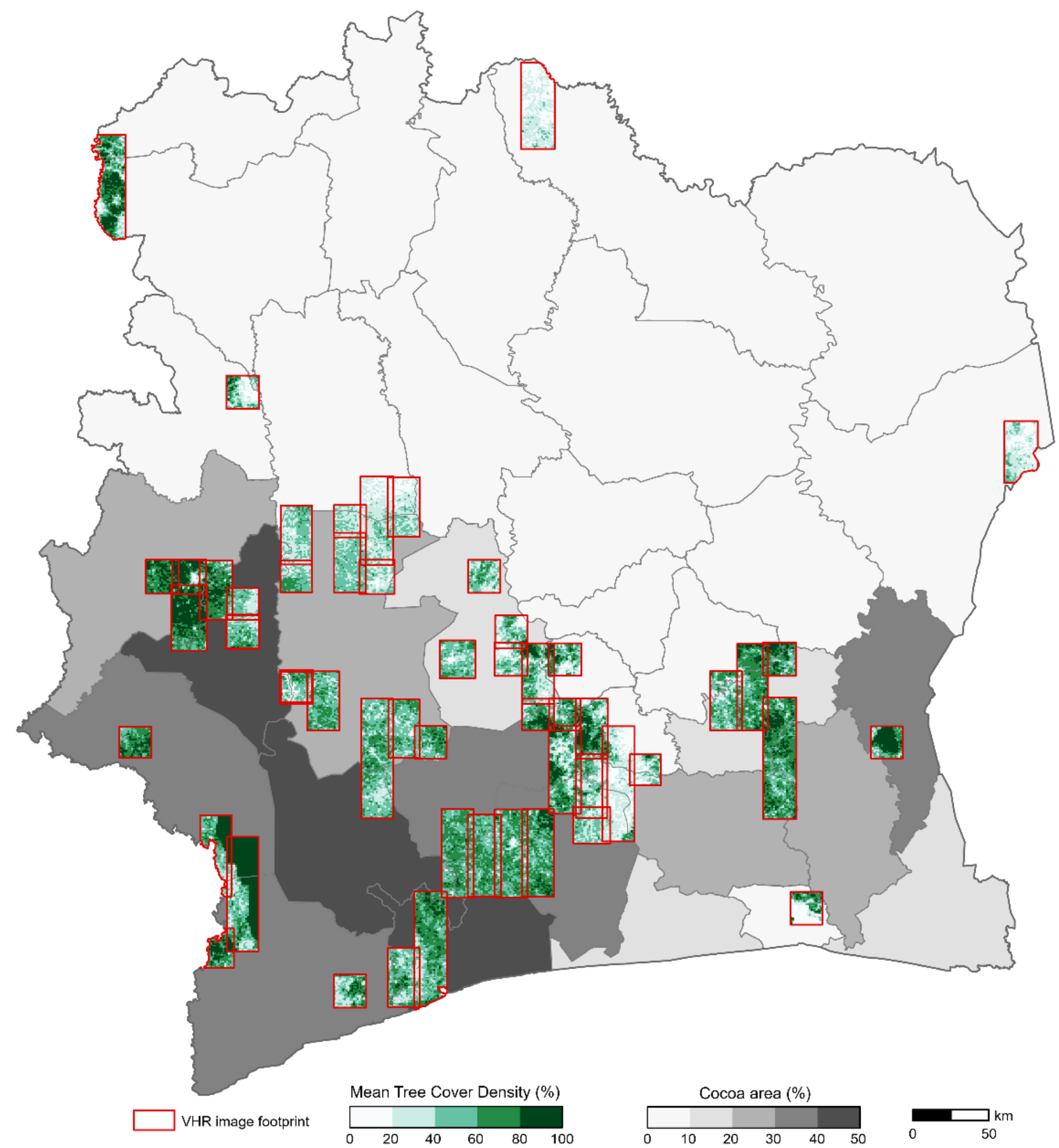

**Figure 2.** Spatial distribution of VHR image footprints defining the study area, and mean Tree Cover Density 2020 from the Copernicus Land Monitoring Service (2025), aggregated to a regular 1 km grid. Cocoa area as a percentage of administrative unit area (level-2) was extracted from the binary cocoa map of Kalischek et al. (2023).

## 2.2 EO datasets and existing mapping products

This study evaluated multiple EO inputs and existing cocoa mapping products, spanning different spatial, spectral, and temporal characteristics, within a unified comparative framework. The evaluated EO inputs comprised VHR optical imagery, a conventional S2 annual composite, and self-supervised foundation-model embeddings derived from multi-sensor EO archives, while existing cocoa mapping products were included as operational benchmarks.

### 2.2.1 Very high resolution imagery

VHR optical satellite imagery from the Pléiades-1A and Pléiades-1B sensors was used as input for VHR model development and inference. The data were made available through the Copernicus Contributing Missions mechanism in the context of the Copernicus Land Monitoring Service (CLMS). The imagery provides four spectral bands (red, green, blue, and near-infrared) at 0.5 m spatial resolution as orthorectified top-of-atmosphere reflectance.

In total, 56 Pléiades scenes acquired between January 2020 and June 2021 were retained after selection based on their geographic coverage of cocoa-producing regions and overlap with available reference data, and after manual screening to exclude scenes affected by extensive cloud cover, severe atmospheric artefacts, or sensor noise. The use of acquisitions from different dates introduces variability in illumination conditions, atmospheric properties, phenological conditions, and viewing geometry, which introduces spectral variability across acquisition conditions but may also improve model robustness to such variation.

The Pléiades data were not acquired specifically for this study but originate from pre-existing acquisitions originally obtained for a previous project (CLS and Terrasphere, 2024). Consequently, the study was limited to areas where suitable VHR imagery and EO-oriented reference data were jointly available and does not provide continuous national coverage.

### 2.2.2 Sentinel-2 composite

A S2 annual composite (2020) was generated in Google Earth Engine using the pan-tropical cloud-masking and compositing approach described by Simonetti et al. (2021). The workflow applies cloud- and cloud-shadow masking procedures developed for tropical regions and computes annual per-band median composites from S2 Level-1C top-of-atmosphere reflectance data using a dedicated tropical cloud and shadow mask. Ten S2 spectral bands were used, including visible (B02–B04), red-edge (B05–B07, B8A), near-infrared (B08), and shortwave infrared bands (B11–B12), at 10 m spatial resolution. This compositing approach was selected because it provides spatially consistent coverage with reduced cloud contamination and atmospheric artefacts in humid tropical environments, and has been specifically designed and validated for land cover mapping applications in tropical regions (Simonetti et al., 2021), avoiding the need for custom compositing procedures. The 2020 target year was selected to align with the majority of comparison products and to ensure consistency with the other EO inputs used in this study. Although the VHR imagery spans January 2020 to June 2021, exact temporal alignment between the VHR imagery and S2 data is inherently limited because VHR imagery consists of single-date acquisitions, whereas annual S2 composites integrate observations acquired throughout the year.

### 2.2.3 Foundation model embeddings

Self-supervised geospatial foundation-model embeddings were evaluated as alternative EO inputs to conventional S2 composites. These approaches compress multi-sensor EO information into dense latent representations intended to capture spectral and temporal patterns while reducing the need for task-specific feature engineering.

TESSERA embeddings (Feng et al., 2025) were generated from multi-temporal S1 and S2 observations using a self-supervised representation-learning framework trained on large-scale EO archives. The resulting embeddings provide 128-dimensional feature representations at 10 m spatial resolution and encode spectral and temporal patterns from the input observations. Annual TESSERA embeddings for 2020 were retrieved for the study area using the geotessera Python package (Feng et al., 2025). TESSERA embeddings were not available as precomputed global archives at the time of this study and were generated on request for the study area.

AEF embeddings were developed by assimilating a diverse range of data sources, including optical imagery (S2, Landsat), radar (S1, PALSAR-2), spaceborne LiDAR (GEDI), terrain (GLO-30), climate reanalysis (ERA5-Land), and geotagged text from species occurrence records and geographic encyclopaedic sources (Brown et al., 2025). In contrast to TESSERA, which generates strictly pixel-wise embeddings with no explicit spatial context (Feng et al., 2025), AEF employs a patch-based architecture using spatial self-attention and convolutional operators, such that each pixel's embedding encodes information from its surrounding spatial neighbourhood (Brown et al., 2025). In this study, 64-dimensional AEF embeddings for 2020 were obtained from Google Earth Engine.

TESSERA and AEF were selected as they represent complementary foundation-model approaches: TESSERA is based on temporal Sentinel observations, whereas AEF is based on broader multi-modal EO inputs. Both provide 10 m spatial resolution embeddings compatible with downstream EO classification workflows.

### 2.2.4 Existing cocoa mapping products

Four publicly available cocoa mapping products were selected for comparison based on their geographic relevance to Côte d’Ivoire, temporal alignment with the study period (2020–2021), methodological transparency, and public accessibility. The selected products represent different EO data sources and modelling approaches currently used for regional or national-scale cocoa mapping. Key characteristics of the products are summarised in Table 1.

**Table 1.** Overview of the publicly available cocoa mapping products used for comparison, including spatial resolution, target year, input data, modelling approach, training data sources, and data access.

| Product | Resolution | Year | Input data | Method | Training data (cocoa) | Access |
|---|---|---|---|---|---|---|
| BNETD (BNETD-CIGN, 2024) | 10 m | 2020 | Sentinel-2 annual mosaic | Random Forest | Field data, Côte d'Ivoire | Google Earth Engine Publisher catalog |
| Kalischek (Kalischek et al., 2023) | 10 m | 2019-2021 | Sentinel-2 time series, Canopy height | Deep learning ensemble | Private sector farm polygons, Côte d'Ivoire and Ghana | Google Earth Engine Publisher catalog |
| FDaP v2025a (Forest Data Partnership, 2025) | 10 m | 2020 | Sentinel-1, Sentinel-2, ALOS PALSAR-2, terrain variables | Deep neural network | Multiple open and proprietary sources | Google Earth Engine Publisher catalog |
| FDaP v2025b (Forest Data Partnership, 2026) | 10 m | 2020 | AlphaEarth Foundations | Deep neural network | Multiple open and proprietary sources | Google Earth Engine Publisher catalog |

## 2.3 Training and validation data

In this study, the reference data used for training and validation were obtained from the COLD-CI dataset described in Orlowski et al. (2026), which provides spatially explicit labels for cocoa and non-cocoa land cover across cocoa-producing regions and contrasting non-cocoa landscapes in Côte d'Ivoire. The dataset was specifically developed for EO-based cocoa mapping applications and was used as the common reference source for all internally trained models.

COLD-CI was generated through the integration of multiple sources, including the West Africa Cocoa (WAC) polygon dataset (Schneider et al., 2023), external thematic datasets, and manual photointerpretation using VHR imagery. To improve label reliability, automated filtering procedures were combined with systematic visual correction and validation.

In COLD-CI, cocoa labels represent cocoa planted area and associated fine-scale internal heterogeneity rather than pure cocoa canopy or complete farm boundaries. This includes cocoa trees and co-occurring non-cocoa trees within agroforestry systems, as well as small non-woody gaps such as open space, herbaceous vegetation, or annual crops occurring within mixed-cropping systems. All non-cocoa land cover types are grouped into a single background class.

The COLD-CI polygon labels were rasterised to the VHR grid using GDAL, with each pixel assigned the class of the polygon whose boundary contained its centre point. Pixels not covered by any polygon were assigned the unlabelled class, as were pixels affected by clouds, cloud shadows, or no-data areas at scene borders in the VHR imagery.

A common tile grid was then used to extract spatially corresponding samples from all EO inputs used to train the internal models and from the rasterised reference labels. A regular grid of 256 × 256 m tiles was generated over the full extent of the VHR imagery and intersected with the rasterised label mask, yielding 176,935 tiles with labelled pixels available for model development. This tile size corresponds to 512 × 512 pixels for the 0.5 m VHR imagery and approximately 26 × 26 pixels for the 10 m inputs. The common tile gridding provided the spatial support for extracting aligned VHR, S2, TESSERA, and AEF inputs.

To create training and validation sets with comparable label and canopy-structure distributions, we applied the Twinning method (Vakayil and Joseph, 2022). The splitting procedure used three tile-level variables: (i) the proportion of labelled pixels, (ii) the proportion of cocoa pixels within labelled areas, and (iii) the proportion of cocoa pixels with overstory cover.

The latter variable was estimated from a 0.5 m canopy height model derived from the VHR imagery using the canopy height estimation approach of Tolan et al. (2024) with pixels exceeding 8 m in height used as a proxy for cocoa areas with overstory cover. Further information on threshold selection can be found in Supplementary Material S1. The dataset was split into approximately 80% training and 20% validation tiles.

## 2.4 Test data

An independent evaluation dataset was collected separately from the COLD-CI reference data used for model development. This design was adopted for three reasons. First, an independent evaluation framework enables a fair comparison between the internally trained models and existing cocoa mapping products. Second, the study required a stratified sampling design to ensure adequate representation of contrasting landscape conditions and sufficient sample sizes for statistically robust stratum-level accuracy estimation. Third, extending the sampling frame beyond the spatial coverage of COLD-CI allowed the full extent of the available VHR imagery footprint to be incorporated into the evaluation. Consequently, test data were collected through a stratified reference-data acquisition procedure.

### 2.4.1 Stratification framework

We designed a stratified evaluation framework to quantify how performance varies across contrasting landscape conditions for internally trained models based on different EO inputs and for existing cocoa mapping products. We hypothesise that signal purity at the pixel level is an important factor influencing the accuracy of medium-resolution cocoa mapping products and EO inputs. Here, signal purity refers to the degree to which the observed EO signal is dominated by cocoa-related information rather than mixed with other land cover components.

As proxies of signal purity, we selected two landscape variables derived from ancillary EO datasets: Tree Cover Density derived from the Copernicus Land Cover and Forest Monitoring (LCFM) Tree Cover Density 2020 (TCD) product (Copernicus Land Monitoring Service, 2025), and the Patch Density (PD) computed from the BNETD Land Cover 2020 dataset (BNETD-CIGN, 2024). Both variables refer to the reference year 2020 and have a 10 m spatial resolution. TCD measures woody tree canopy closure, including cocoa trees and overstory shade trees but excluding shrubs, on a continuous scale from 0–100%. Based on qualitative assessment of cocoa-producing areas covered by the VHR imagery, full-sun cocoa plantations generally exhibited TCD values below 70%. TCD values exceeding this threshold are therefore indicative of substantial overstory shade tree presence. Below this threshold, TCD reflects variation in cocoa canopy development and density across full-sun and lightly shaded production systems. TCD therefore serves as a proxy for canopy conditions expected to influence cocoa signal purity across the full range of cocoa production systems. Three broad TCD regimes were identified based on this assessment:

- **TCD 0–30%:** Areas ranging from very young to sparsely planted cocoa systems, where spectral responses are expected to be dominated by non-tree components such as soil or herbaceous vegetation.
- **TCD 30–70%:** Areas with partial to high canopy closure, typically corresponding to developing or moderately dense to mature cocoa plantations, or to agroforestry systems with scattered and mostly isolated shade trees. These conditions are expected to produce partially mixed spectral signal, although the cocoa signal prevails.
- **TCD >70%:** Areas with high canopy closure, often corresponding to cocoa systems with moderate to dense overstory cover. Under these conditions, overstory vegetation may dominate the spectral response, partially or fully obscuring the cocoa signal. High canopy cover in this stratum typically originates from mixed forest–cocoa landscapes or from cocoa intercropped with taller tree crops such as rubber or oil palm.

PD was introduced as a second stratification variable to capture landscape fragmentation. Here, fragmentation refers to the degree of spatial patchiness within the landscape and is expected to influence the extent of spectral mixing along land-cover boundaries. Highly fragmented landscapes contain a larger proportion of edge pixels and small land cover patches, reducing signal purity. Small or narrow cocoa patches may additionally fall below the effective spatial support of medium-resolution EO products such as S2, further compromising the cocoa signal. PD measures the number of land cover patches per 100 ha, following the convention established in landscape ecology (McGarigal and Marks, 1995). It was calculated using all classes in the original BNETD Land Cover 2020 dataset. Patches were defined as contiguous groups of pixels belonging to the same class using an 8-neighbour connectivity rule. The global PD distribution was divided into tertiles representing low, medium, and high fragmentation conditions.

To analyse the impact of these characteristics on classification accuracy, we stratified all tiles within the full extent of the VHR imagery while excluding tiles used for model training and validation. Tiles located within overlapping VHR scene areas were additionally removed to prevent multiple representations of the same geographic locations. After applying these exclusions, 343,019 tiles remained available for strata assignment and evaluation.

Each tile was assigned to the dominant TCD class within the tile extent, determined from the modal TCD value, and to a PD class based on its position within the global tile-level PD tercile distribution.

Finally, TCD and PD classes were crossed to construct a two-dimensional stratification framework consisting of 9 landscape strata.

### 2.4.2 Test data collection

Since the stratified test tile pool was purposively extended beyond the spatial coverage of the COLD-CI reference dataset, reference labels were obtained through independent expert photointerpretation of VHR imagery.

Within each stratum, reference points were randomly distributed across the available tiles, with one random location generated within each sampled tile. The required sample size was estimated using the worst-case formula for estimating a proportion, $n = z^2 \times p(1 - p) / e^2$, where $n$ is the required sample size, $z = 1.96$ corresponds to a 95% confidence level, $p$ is the expected proportion, and $e = 0.05$ represents the target margin of error ($\pm 5\%$). The maximum variance, $p(1 - p) = 0.25$, occurs at $p = 0.5$, yielding approximately 384 points per stratum. Interpretation proceeded iteratively using randomly drawn points until the Wilson confidence-interval half-width criterion ($\pm 5\%$) was met within a stratum or 384 points had been interpreted, whichever occurred first. This ensured that sampling continued until the slowest-converging product satisfied the precision target.

Finally, per-stratum sample sizes ranged from 189 to 374 points (total n = 2,821; see Supplementary Material S2, Table S1 for stratum-level sample sizes and confidence-interval half-widths).

## 2.5 Model development

Internally trained models were developed using identical reference data, training-validation splits, model architecture, and optimisation procedures to enable controlled comparison of EO inputs independent of differences in training data or model design.

### 2.5.1 Network architecture

We used a U-Net convolutional neural network architecture (Ronneberger et al., 2015) with an EfficientNet-B5 encoder (Tan and Le, 2019). EfficientNet employs a compound scaling strategy that balances network width, depth, and resolution, enabling efficient extraction of multi-scale feature representations.

### 2.5.2 Model training

For each EO input, an ensemble of five models was trained using identical network architectures but different random initialisations, data augmentation realisations, and batch sampling. For the VHR ensemble, the input consisted of four spectral bands (red, green, blue, and near-infrared). S2 ensembles used the ten-band annual composite described in Section 2.2.2, while the TESSERA and AEF ensembles used the corresponding embedding dimensions as input features. All input channels were standardised using channel-specific mean and standard deviation values computed across all training tiles. Data augmentation included random flips and rotations applied with a probability of 0.5, consistent across all EO inputs to maintain identical training conditions. Radiometric augmentations were not applied because the foundation-model embedding inputs represent learned feature embeddings rather than physical spectral measurements.

The 10 m S2 and foundation-model inputs were zero-padded to 32 × 32 pixels prior to model input to satisfy the minimum spatial input requirements of the EfficientNet-B5 encoder while preserving the original spatial information content. The VHR inputs retained their native 512 × 512 pixel dimensions.

Models were trained using a batch size of 16, a learning rate of $1 \times 10^{-3}$, and a maximum of 80 epochs with early stopping based on validation loss using a patience of five epochs. A cross-entropy loss function was used with unlabelled pixels excluded from loss computation.

Final predictions were generated by averaging class probabilities across ensemble members to improve predictive robustness and reduce sensitivity to individual model initialisation and sampling variability.

### 2.5.3 Spectral ablation experiment

We hypothesise that part of the performance difference between the S2 and VHR models reflects the balance between two opposing characteristics: lower spatial resolution, which increases the accuracy gap, and richer spectral content, which partially offsets it.

To assess this hypothesis, an additional S2 model (S2_4b) was trained using only the four spectral bands which were also available in the Pléiades imagery (red, green, blue, and near-infrared). All other aspects of model development, including training and validation data, data splits, model architecture, and optimisation procedure, were identical to those used for the other internally trained models. Comparing S2_4b with the VHR model provides a band-matched comparison that helps assess the contribution of spatial resolution, although other sensor and acquisition differences cannot be fully excluded.

## 2.6 Accuracy metrics and uncertainty estimation

For the stratified evaluation, ensemble-averaged probability outputs of the internally trained models were thresholded at $p \geq 0.5$ to generate binary cocoa/non-cocoa predictions. Existing cocoa mapping products delivering continuous probability outputs were binarised prior to evaluation using thresholds reported by the original authors: Kalischek at $p \geq 0.65$ (Kalischek et al., 2023), FDaP v2025a at $p \geq 0.96$ (Forest Data Partnership, 2025), and FDaP v2025b at $p \geq 0.5$ (Forest Data Partnership, 2026).

Model and product performance was evaluated using F1-score, user's accuracy (UA), producer's accuracy (PA), and overall accuracy (OA). UA and PA for the cocoa class correspond to precision and recall, respectively, while the F1-score represents their harmonic mean.

Because points were allocated equally across strata rather than proportionally to stratum area, stratum-level confusion matrices were aggregated using area-based weights following Olofsson et al. (2014), where each stratum weight reflected its proportional share of the total tile count; since all tiles were of uniform size, this is equivalent to weighting by relative stratum area. OA, UA, PA and F1-score were subsequently derived from the resulting area-adjusted confusion matrix.

95% confidence intervals were estimated for OA following Olofsson et al. (2014) using the variance of the area-adjusted accuracy estimator. FDaP v2025b (Forest Data Partnership, 2026) was released and included in the evaluation after the reference data collection had been completed. Its accuracy estimates are therefore based on the existing reference points without additional sampling. As a consequence, the $\pm 5\%$ precision criterion for stratum-level OA was not met for this product in three strata.

# 3 Results

The stratified evaluation revealed substantial variation in cocoa mapping performance across landscape conditions. Figure 3 illustrates representative cocoa probability maps from the four internally trained models based on different EO inputs, highlighting differences in prediction patterns between VHR and medium-resolution approaches. While the general spatial distribution of cocoa probabilities was often similar across models, the VHR model preserved substantially greater fine-scale spatial variation and local detail, whereas the medium-resolution models produced more spatially aggregated prediction surfaces.

As shown in Figure 4, the observed performance differences were strongly associated with both TCD and landscape fragmentation. Across all internally trained models and existing cocoa mapping products, the highest F1-scores were observed within the medium-TCD regime (30–70%) under low-fragmentation conditions. Performance declined toward highly fragmented landscapes and under very low or very high TCD conditions. These declines were particularly pronounced for the S2 model, BNETD, and the FDaP products, whereas the VHR model exhibited comparatively limited sensitivity to these landscape gradients and maintained consistently high performance across all strata. Complete per-stratum accuracy metrics for all evaluated models and products are provided in Supplementary Material S3, Table S2.

Among the internally trained models (Figure 4), the VHR model achieved the highest and most consistent performance, with F1-scores ranging from 0.90 to 0.96 across strata and peaking in the medium-TCD/low-fragmentation stratum. TESSERA performed closest to the VHR model, with F1-scores ranging from 0.83 to 0.93 and F1 differences relative to VHR between 0.03 and 0.08. AEF showed intermediate performance between TESSERA and S2 across all strata, while the S2 baseline consistently produced the lowest performance among the internally trained models.

The spectral ablation experiment (Supplementary Material S4, Table S3) showed that the S2_4b model achieved an F1-score of 0.64 and OA of 0.76, compared with 0.76 and 0.85, respectively, for the full ten-band S2 model. The VHR model (F1 = 0.92) outperformed S2_4b by 0.28 F1 units, a substantially larger margin than the 0.16 F1-unit gap observed between VHR and the full ten-band S2 model. This indicates that the additional S2 spectral bands recovered approximately 0.12 F1 units, or around 43% of the band-matched VHR–S2_4b performance gap.

The existing cocoa mapping products showed larger performance variability across landscape conditions (Figure 4). Kalischek achieved the strongest overall performance among the external products, with F1-scores ranging from 0.69 to 0.91. Notably, in the medium-TCD/low-fragmentation stratum, Kalischek achieved an F1-score of 0.91, exceeding the internally trained S2 model (0.87) within the same stratum. In contrast, BNETD exhibited substantially lower performance under highly fragmented conditions, with F1-scores decreasing to 0.38–0.49 in the high-fragmentation strata. FDaP v2025a and FDaP v2025b showed intermediate performance across most landscape conditions. FDaP v2025a achieved higher F1-scores in the low-TCD strata, whereas FDaP v2025b matched or slightly exceeded FDaP v2025a under medium- and high-TCD conditions.

Following the stratified analysis, Table 2 summarises area-weighted OA, UA, PA, and F1-score for all internally trained models and existing cocoa mapping products. The weighted estimates broadly confirmed the stratified results. The VHR model achieved the highest weighted OA (0.94) and F1-score (0.92), followed by TESSERA (OA = 0.90; F1 = 0.86). Among the externally developed products, Kalischek et al. (2023) achieved the highest F1-score (0.83), whereas BNETD achieved the lowest (0.55). OA confidence-interval half-widths ranged from 0.010 to 0.021, indicating comparatively low sampling uncertainty relative to the observed performance differences between products. The wider confidence interval for FDaP v2025b reflects its inclusion after completion of the reference-data collection (see Section 2.6).

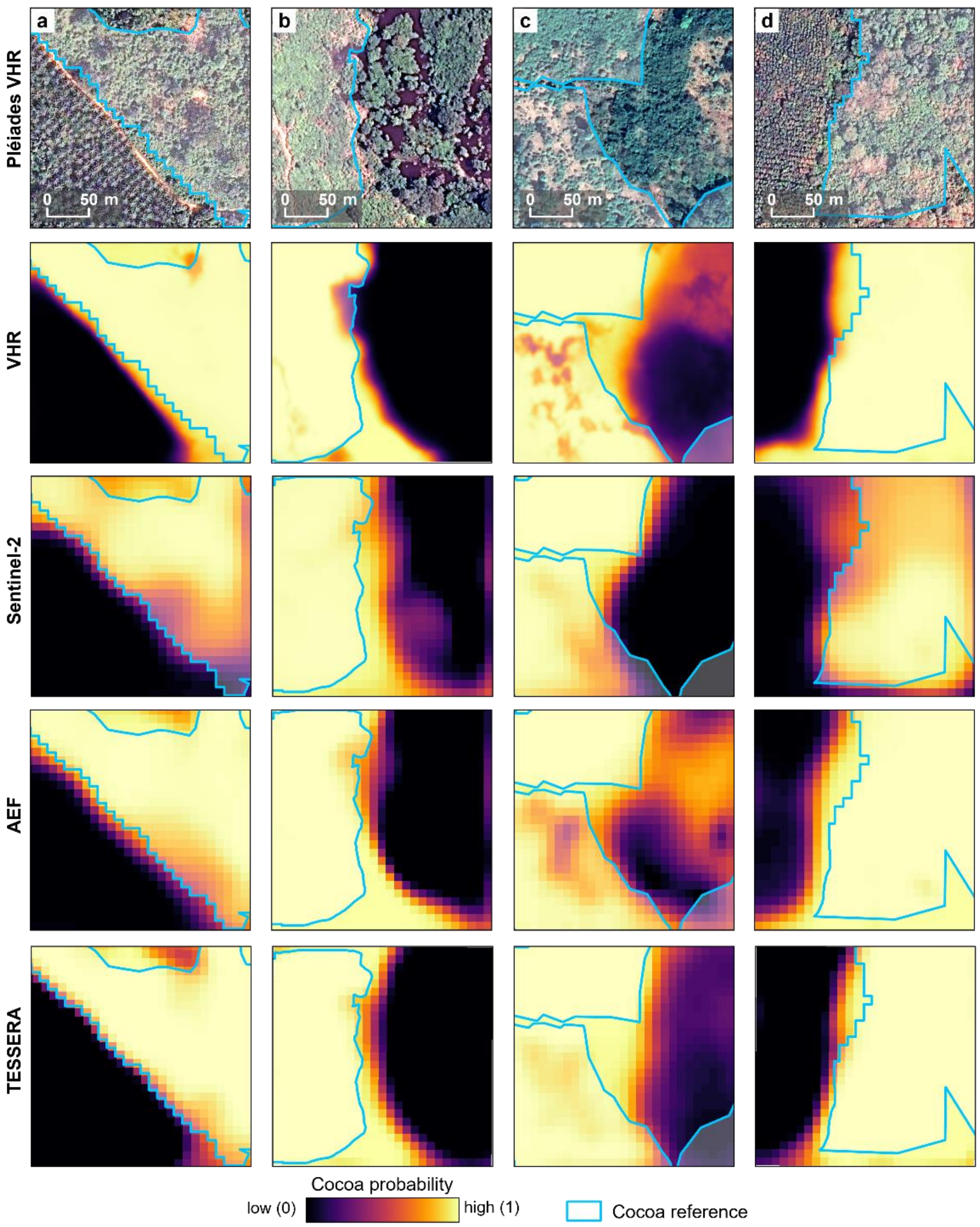

**Figure 3.** Examples of cocoa probability predictions across different landscape types for the evaluated EO inputs. The first row shows the corresponding 0.5 m Pléiades very high resolution (VHR) imagery. Rows 2–5 show cocoa probability predictions generated using the VHR, Sentinel-2, AlphaEarth Foundations (AEF), and TESSERA EO inputs, respectively. Cyan outlines indicate the cocoa reference polygons from the COLD-CI dataset (Orlowski et al., 2026). Panels illustrate representative landscape situations: (a) cocoa plantation bordering oil palm, (b) cocoa plantation bordering forest, (c) sparse cocoa planting with heterogeneous canopy cover, and (d) cocoa agroforestry system bordering a rubber plantation. VHR image data: © CNES (2020), distribution Airbus Defence and Space.

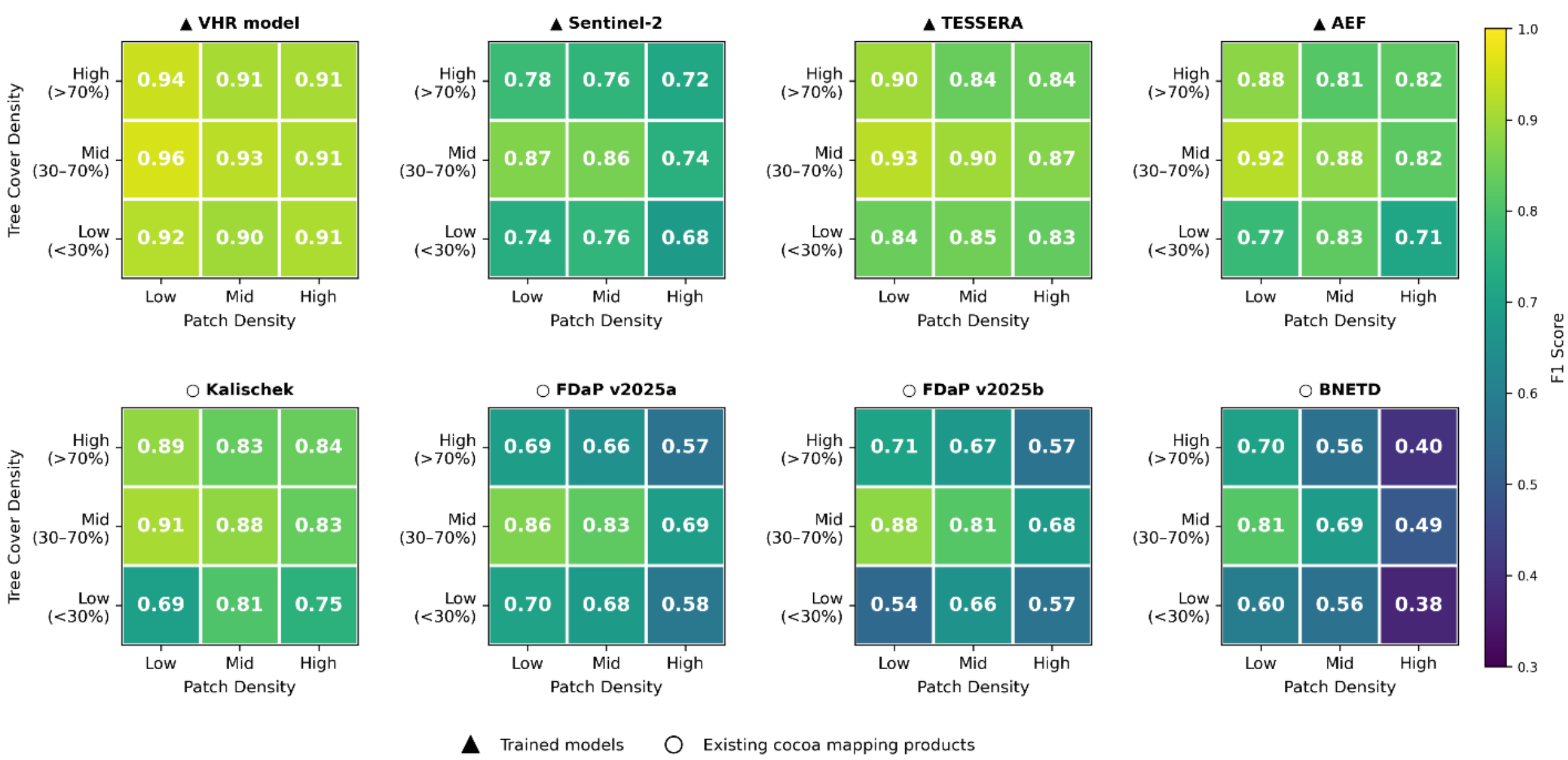

**Figure 4.** Stratified F1-score comparison across tree cover density (TCD) and landscape fragmentation gradients for the internally trained models and existing cocoa mapping products. Rows represent TCD classes derived from the Copernicus LCFM Tree Cover Density 2020 product, while columns represent Patch Density (PD) fragmentation classes derived from the BNETD Land Cover 2020 dataset. Cell values indicate F1-scores computed from the independent stratified random reference sample. Internally trained models are marked with triangles (▲), while existing cocoa mapping products are indicated by circles (○). Higher F1-scores indicate better classification performance.

**Table 2.** Accuracy metrics derived from the area-adjusted confusion matrix based on the stratified random sample of reference points. OA = area-weighted overall accuracy; F1 = F1-score; UA = user's accuracy for the cocoa class; PA = producer's accuracy for the cocoa class; CI95% = half-width of the 95% confidence interval on the area-weighted OA.

| EO representation/Product | F1 | OA | UA | PA | CI95% |
|---|---|---|---|---|---|
| VHR | 0.92 | 0.94 | 0.85 | 1.00 | 0.010 |
| TESSERA | 0.86 | 0.90 | 0.79 | 0.93 | 0.013 |
| AEF | 0.82 | 0.87 | 0.74 | 0.92 | 0.014 |
| Kalischek | 0.83 | 0.90 | 0.89 | 0.77 | 0.013 |
| Sentinel-2 | 0.76 | 0.85 | 0.76 | 0.75 | 0.016 |
| FDaP v2025a | 0.66 | 0.72 | 0.54 | 0.84 | 0.019 |
| FDaP v2025b | 0.64 | 0.67 | 0.49 | 0.91 | 0.021 |
| BNETD | 0.55 | 0.74 | 0.63 | 0.49 | 0.019 |

# 4 Discussion

## 4.1 Very high resolution imagery provides measurable and consistent advantages for cocoa mapping

The VHR model consistently achieved the highest accuracy among all internally trained models and existing cocoa mapping products. The advantage of VHR imagery was not limited to overall performance metrics but was observed across all landscape strata, where the VHR model maintained F1-scores above 0.90 and exhibited substantially lower sensitivity to variation in TCD and landscape fragmentation than any of the 10 m approaches. These results indicate that the additional spatial detail provided by sub-metre imagery translates into more robust cocoa detection across a wide range of landscape conditions.

In Côte d'Ivoire, cocoa is predominantly cultivated by smallholders within heterogeneous agricultural mosaics containing spatial features considerably smaller than a 10 m pixel, as directly observable in the VHR imagery used in this study. Under such conditions, pixels located within cocoa planted areas may contain spectral contributions from cocoa trees, co-occurring non-cocoa trees, and small non-woody gaps such as open space, herbaceous vegetation, or annual crops. Near planted-area boundaries, pixels may additionally include contributions from adjacent land-cover types. Although this fine-scale internal heterogeneity is included within the COLD-CI definition of cocoa planted area, it can reduce the distinctiveness of the cocoa signal at medium spatial resolution. VHR imagery reduces this spatial aggregation by resolving many of these components directly, thereby enabling more precise delineation of cocoa planted-area boundaries and reducing ambiguity in the observed EO signal.

Under highly fragmented conditions, the medium-resolution approaches experienced notable performance declines, while the VHR model maintained consistently high accuracy. Because architecture, training data, data splits, and optimisation procedures were held constant across the internally trained models, the observed differences are primarily attributable to the characteristics of the EO inputs themselves. The results therefore suggest that, in fragmented cocoa landscapes, spatial resolution represents an important asset that cannot be fully compensated by additional spectral or temporal information. Once multiple land-cover components are aggregated within a single medium-resolution pixel, some of the fine-scale information relevant to cocoa identification is no longer retained and cannot be fully recovered through richer medium-resolution representations.

The spectral ablation experiment helps disentangle the contribution of spectral information to the observed VHR–S2 performance difference (Supplementary Material S4, Table S3). When S2 was restricted to the four bands shared with the VHR imagery, the F1 gap widened to 0.28, compared with 0.16 for the full ten-band S2 model. Restoring the additional six S2 bands therefore recovered approximately 43% of the band-matched gap. However, despite its richer spectral information, the full S2 model still remained substantially below the VHR model. This suggests that the additional spectral information available at 10 m cannot fully compensate for the finer spatial detail provided by the VHR imagery, although other sensor and acquisition differences cannot be fully excluded.

## 4.2 Performance varies more across tree cover density than fragmentation

Across the two stratification dimensions, TCD emerged as the stronger driver of cocoa mapping performance for medium-resolution EO inputs and existing cocoa mapping products, with performance differences across TCD strata generally exceeding those observed across fragmentation levels. All evaluated approaches achieved their highest accuracies within the intermediate TCD range (30–70%), which corresponds to cocoa production systems with moderate to high canopy closure, including mature full-sun plantations and agroforestry systems with scattered shade trees, where the cocoa signal prevails. In contrast, performance declined under both low and high TCD conditions. For medium-resolution EO inputs, this pattern is consistent with the signal-purity hypothesis proposed in Section 2.4.1. In low-TCD cocoa areas, sparse or incompletely closed canopies increase contributions from soil and herbaceous vegetation, weakening the cocoa signal. At TCD values above 70%, dense canopy cover is not unique to cocoa: within plantations, overstory shade may obscure the cocoa layer and increase omission errors, whereas forests and other perennial tree crops, such as rubber or cashew, may be confused with cocoa and increase commission errors. Performance declines at high TCD therefore likely reflect both canopy masking and reduced separability from other densely wooded land-cover types.

Variation in VHR model performance across TCD strata followed a similar pattern, but with smaller performance reductions outside the intermediate TCD range. This may be explained by different underlying mechanisms. At sub-metre resolution, pixel-level mixing becomes substantially less important, as individual canopy elements, gaps, and spatial structures are directly observable. Instead, performance appears to depend on the visibility and identifiability of cocoa canopies. Under low-TCD conditions, cocoa trees may be fully visible but exhibit limited canopy development, making them more difficult to distinguish from other tree crops or woody vegetation. Conversely, under high-TCD conditions, dense overstory vegetation may partially or completely obscure the cocoa layer. In such situations, increasing spatial resolution alone cannot fully overcome the limitations imposed by canopy occlusion, as the cocoa trees may become increasingly difficult to observe directly.

At the same time, VHR imagery provides contextual information that extends beyond the cocoa canopy itself and that can be leveraged by spatially aware models. Planting patterns, crown spacing, canopy texture, field boundaries, access paths, small clearings, and built structures associated with agricultural activities, when present within the context window (a tile), may help distinguish cocoa production systems from surrounding land cover even when cocoa trees are only partially visible. Such contextual information is largely unavailable at 10 m resolution and may contribute to the comparatively stable performance of the VHR model across contrasting landscape conditions. Collectively, these findings suggest that the superior performance of VHR imagery arises not only from its ability to resolve individual cocoa trees, but also from its capacity to capture the broader spatial context in which cocoa cultivation occurs.

## 4.3 Foundation-model embeddings improve cocoa mapping performance but do not eliminate the spatial-resolution gap

Among the internally trained 10 m models, both TESSERA (Feng et al., 2025) and AEF (Brown et al., 2025) consistently outperformed the S2 baseline, indicating that foundation-model embeddings capture information relevant to cocoa detection. TESSERA achieved the strongest performance among the evaluated 10 m inputs and substantially reduced the performance gap relative to the VHR model. While the present study was not designed to identify the specific causes of this difference, the results indicate that the choice of foundation-model embedding can have a measurable impact on cocoa mapping performance.

This result is notable given the contrasting designs of the two foundation models. TESSERA generates pixel-wise embeddings without explicit spatial-context encoding and represents temporal dynamics using S1 and S2 observations alone (Feng et al., 2025). In contrast, AEF processes spatial patches using self-attention and convolutional operators, encoding both temporal dynamics and spatial neighbourhood information while additionally assimilating a substantially broader range of data sources (Brown et al., 2025). Despite this broader input set and explicit spatial-context encoding, AEF consistently underperformed TESSERA for cocoa mapping.

A similar pattern was reported by Lisaius et al. (2026a), who found that TESSERA outperformed AEF for crop-type classification in a smallholder agricultural landscape in Senegal, with the gap widening substantially under temporal transfer relative to same-year evaluation. The authors hypothesised that AEF's broader multimodal representation and explicit spatial-context encoding may blur the fine-grained spectral–temporal signals needed to distinguish spectrally similar classes. Our results are compatible with this explanation and suggest that TESSERA's more targeted representation, supported by day-of-year positional encoding and robustness to irregular temporal sampling, may be better suited to discriminating cocoa from similar land-cover types (Feng et al., 2025).

Further support is provided by a separate small-field study in Austria, where TESSERA outperformed both conventional Sentinel inputs and AEF and maintained comparatively strong classification performance near field boundaries, suggesting lower sensitivity to mixed-pixel and edge effects (Lisaius et al., 2026b). Although none of these studies directly isolated the mechanism responsible for TESSERA's advantage, the consistent pattern across cocoa mapping in Côte d'Ivoire, smallholder crop classification in Senegal, and small-field agriculture in Austria suggests that embedding design may influence sensitivity to fine-grained spectral–temporal signals and landscape heterogeneity.

This broader interpretation is also consistent with Ma et al. (2026), who found that AEF embeddings were competitive with purpose-built remote-sensing features across several agricultural applications when models were trained and evaluated locally, but did not consistently outperform them and showed limitations in temporal sensitivity and interpretability. Foundation-model performance therefore depends on embedding design, the downstream task, and the broader modelling and evaluation setting. Broader and more diverse input modalities do not necessarily translate into better performance for a specific mapping application. In the present study, although both TESSERA and AEF improved upon the conventional S2 baseline, neither matched the VHR model, and the performance of both varied across fragmentation and TCD strata.

## 4.4 Comparison of existing cocoa mapping products

The comparison of the existing cocoa mapping products broadens the perspective developed in Section 4.3 by showing that performance depends not only on the EO input or embedding design, but on the complete mapping workflow.

Kalischek et al. (2023) achieved slightly higher overall performance than the internally trained AEF model, although their overall F1-scores were comparable. This is particularly relevant because both approaches were developed specifically for cocoa mapping in West Africa and used deep-learning and ensemble-based workflows. Kalischek et al. combined more than 100,000 georeferenced cocoa farm polygons with explicit multi-temporal S2 observations and canopy-height information. These inputs may retain task-relevant temporal and structural signals that are less directly represented in the compressed AEF embeddings and may therefore have contributed to its strong performance across most landscape strata. However, the small performance difference cannot be attributed to these inputs alone because the two approaches also differ in other aspects of model design and implementation.

In contrast, BNETD was developed as a general national land-cover product rather than a dedicated cocoa mapping system. It relied on a comparatively limited cocoa reference dataset, a single annual S2 composite, and a pixel-wise Random Forest classifier, and therefore did not explicitly incorporate temporal dynamics or spatial context (BNETD-CIGN, 2024). Model development was also affected by temporal discrepancies between field reference-data collection and the target mapping year. Its pronounced performance declines under highly fragmented landscapes and low- and high-TCD conditions are consistent with the limitations of a pixel-wise, single-composite approach where cocoa signals may be spatially mixed, weak, or obscured.

The comparison between FDaP v2025a and FDaP v2025b provides additional evidence that advances in EO inputs alone do not necessarily translate into improved map quality. Despite the introduction of AEF embeddings in FDaP v2025b, overall performance remained broadly comparable to the earlier FDaP v2025a product, which was developed using multi-sensor information (Forest Data Partnership, 2025, 2026). While v2025b achieved higher PA, this gain was accompanied by lower UA, resulting in similar overall F1-scores. It should be noted, however, that the two product versions employ different recommended probability thresholds ($p \geq 0.96$ for v2025a and $p \geq 0.5$ for v2025b), which may contribute to the observed shift in the balance between UA and PA. Nevertheless, the results suggest that the use of foundation-model embeddings alone does not guarantee higher overall mapping performance.

The comparison between the internally trained AEF model and FDaP v2025b further illustrates the importance of factors beyond the EO input alone. Despite both approaches relying on AEF embeddings as their primary EO input (Brown et al., 2025; Forest Data Partnership, 2026), the internally trained AEF model achieved substantially higher performance than FDaP v2025b (F1 = 0.82 vs 0.64). Several factors may contribute to the observed performance gap. First, COLD-CI underwent substantial manual quality control and boundary refinement, providing representative negative samples and consistent annotation procedures explicitly designed to support deep-learning-based cocoa mapping (Orlowski et al., 2026). Second, the internally trained model was developed exclusively for Côte d'Ivoire and may therefore have learned country-specific cocoa spectral and landscape characteristics more effectively than a global model required to generalise across diverse production systems and regions. Third, the combination of spatially informed AEF embeddings and a U-Net architecture capable of learning spatial relationships across the tile may have further amplified spatial discriminability compared with FDaP v2025b's per-pixel approach, although this cannot be directly assessed from the present comparison.

Collectively, these comparisons indicate that cocoa mapping performance depends on the complete mapping workflow, including the EO input, model architecture, the incorporation of spatial context and temporal dynamics, and the size, regional relevance, and temporal alignment of the reference data. However, because these factors differ simultaneously among the evaluated products, the contribution of any individual component cannot be isolated from the present comparison.

## 4.5 Implications for operational cocoa mapping and regulatory monitoring

The results suggest that the optimal choice of cocoa mapping approach depends on the operational context, balancing the spatial scale of analysis, required thematic accuracy, and intended use of the output.

For large-scale baseline mapping, the scalability, temporal coverage, and historical availability of medium-resolution EO data remain major advantages. This is particularly relevant for retrospective analyses, such as the establishment of commodity baselines for historical reference years, where suitable VHR imagery may not be available wall-to-wall. Foundation-model embeddings may offer an attractive compromise for such applications: both TESSERA and AEF substantially improved performance compared to conventional S2 inputs while retaining the spatial coverage and historical archive availability required for wall-to-wall mapping.

Where existing cocoa mapping products are used for large-area applications, their contrasting error profiles also have important operational implications. Kalischek achieved the highest UA among the existing products (0.89) but comparatively lower PA (0.77), indicating relatively few commission errors but omission of a substantial proportion of actual cocoa. Conversely, FDaP v2025b achieved high PA (0.91) but low UA (0.49), indicating broad cocoa detection accompanied by substantial commission error. These contrasting profiles suggest that product suitability depends on the intended application. Products with high UA may be preferable where false-positive cocoa detections are particularly costly, whereas products with high PA may be more appropriate for broad screening when subsequent verification is available.

The role of VHR imagery is different. Rather than serving as the primary data source for national-scale mapping, VHR observations are most valuable where higher thematic accuracy is required or where early detection of new cocoa expansion is critical. The superior performance of the VHR model under low-TCD conditions indicates an improved ability to identify sparse or incompletely closed cocoa plantations that may remain difficult to detect using medium-resolution observations, potentially resulting in a substantial delay between plantation establishment and its detection in operational monitoring systems. Similarly, the strong performance under high-TCD conditions suggests that VHR imagery may improve the identification of cocoa cultivated under relatively dense canopy conditions within Côte d'Ivoire, where medium-resolution observations often struggle to separate cocoa from surrounding woody vegetation. These characteristics may be particularly valuable for monitoring cocoa expansion into natural forests and protected areas, where timely detection can be critical for management and enforcement actions.

These roles need not be viewed as mutually exclusive. Hybrid cross-scale strategies can combine targeted VHR observations with the coverage and temporal depth of medium-resolution data. Liu et al. (2026), for example, used VHR-derived fine-scale information to support large-area mapping with S2 time series. A comparable cocoa-mapping strategy could use targeted VHR coverage to generate or refine training labels in structurally complex landscapes, while medium-resolution models provide spatially continuous mapping over larger areas.

Beyond its role in training-data generation, VHR imagery could also be deployed selectively for operational mapping. Medium-resolution inputs, particularly foundation-model embeddings, could support wall-to-wall baseline mapping and broad deforestation-risk screening, complemented by targeted VHR analysis in landscapes where medium-resolution predictions are less reliable or where timely and precise detection is particularly important.

## 4.6 Limitations

Several limitations should be considered when interpreting the results of this study. First, the spatial extent of the study was constrained by the availability of suitable VHR imagery, with the analysis limited to 56 Pléiades scenes rather than continuous national coverage. Although the selected scenes span a broad range of cocoa-producing landscapes and environmental conditions across Côte d'Ivoire, they may not capture the full range of cocoa production systems and landscape conditions present in the country. Consequently, the reported accuracy estimates should not be interpreted as nationally representative. Moreover, mean shade-tree cover across cocoa-growing areas in Côte d'Ivoire is relatively low (Becker et al., 2025). The relative advantage of VHR observed under the densest canopy conditions represented in this study may therefore differ in other cocoa-producing regions where agroforestry systems have higher shade-tree cover and greater structural complexity. Nevertheless, the mechanisms examined here, including variation in canopy conditions, cocoa canopy development, and landscape fragmentation, are likely to remain relevant beyond Côte d'Ivoire, although the magnitude of their effects may differ across cocoa-producing regions. Further evaluation is needed to determine whether the observed performance patterns extend to more structurally complex agroforestry systems.

Second, the spectral ablation experiment helped distinguish the contribution of the six additional S2 bands, but the comparison could not fully isolate spatial resolution from other differences between the VHR and S2 inputs, including sensor characteristics and acquisition dates. The remaining performance gap should therefore be interpreted as evidence that finer spatial resolution is likely the primary contributor, rather than as a precise estimate of the effect of spatial resolution alone.

Finally, a common probability threshold ($p \geq 0.5$) was applied to the outputs of all internally trained models rather than optimising thresholds individually for each model. While this ensured a consistent comparison framework, optimising the threshold separately for the ensemble-averaged probability output of each EO input could improve performance to varying degrees and therefore alter the magnitude of the reported differences.

# 5 Conclusion

This study evaluated whether sub-metre imagery provides meaningful advantages over decametric EO inputs for cocoa mapping in Côte d'Ivoire. The VHR model achieved the highest performance (F1 = 0.92) and maintained F1-scores above 0.90 across all evaluated landscape strata. The magnitude of its advantage depended on landscape context and was most pronounced under conditions associated with greater mapping difficulty, including low- and high-TCD conditions and highly fragmented landscapes.

Foundation-model embeddings substantially improved decametric cocoa mapping, with TESSERA achieving the strongest performance among the 10 m inputs (F1 = 0.86). However, they did not eliminate the performance gap to VHR or the landscape-dependent declines observed across the medium-resolution inputs. These patterns suggest that fine-scale spatial information lost through spatial aggregation cannot be fully compensated for by more sophisticated EO representations alone.

By accounting for landscape diversity across cocoa-producing areas in Côte d'Ivoire, the stratified evaluation framework revealed performance variations that were otherwise obscured by metrics aggregated across the full study area. These findings support the use of stratified accuracy assessments when evaluating cocoa mapping products across diverse cultivation systems, ranging from young or sparsely planted cocoa to mature full-sun monocultures and dense agroforestry systems, and across landscapes with varying levels of fragmentation.

Overall, our findings suggest that VHR-based cocoa mapping may be most suitable where the highest possible thematic accuracy is required, such as in regulatory contexts where mapping errors can carry significant consequences. For large-area baseline mapping, foundation-model embeddings offer a compelling balance of accuracy and operational scalability. Operational systems may therefore benefit from combining medium-resolution embeddings for broad-area mapping with targeted VHR observations in locations characterised by high uncertainty or particular regulatory importance.

# CRediT authorship contribution statement

**Kasimir Orlowski:** Conceptualization, Methodology, Software, Formal analysis, Investigation, Data curation, Visualization, Writing – original draft, Writing – review and editing. **Filip Sabo:** Methodology, Software, Investigation, Data curation, Writing – review and editing. **Michele Meroni:** Conceptualization, Methodology, Supervision, Writing – review and editing. **Astrid Verhegghen:** Conceptualization, Writing – review and editing. **Mariana Belgiu:** Supervision, Writing – review and editing. **Felix Rembold:** Project administration, Supervision.

# Funding

This research did not receive any specific grant from funding agencies in the public, commercial, or not-for-profit sectors.

# Data statement

The COLD-CI reference data used for model development are described in Orlowski et al. (2026), which provides the corresponding data-access information. The validation data collected for this study are available from the corresponding author upon reasonable request. The Pléiades imagery cannot be redistributed because of licensing restrictions.

# Acknowledgements

Very high resolution imagery used in this study was made available through the Copernicus Contributing Missions mechanism in the context of the Copernicus4GEOGLAM activities of the Copernicus Land Monitoring Service (CLMS).

# References

Abu, I.O., Szantoi, Z., Brink, A., Robuchon, M., Thiel, M., 2021. Detecting cocoa plantations in Côte d'Ivoire and Ghana and their implications on protected areas. Ecol. Indic. 129, 107863. https://doi.org/10.1016/j.ecolind.2021.107863

Ball, J.G., Wicklein, J.A., Feng, Z., Knezevic, J., Jaffer, S., Madhavapeddy, A., Atzberger, C., Dalponte, M., Coomes, D., 2026. Geospatial foundation models enable data-efficient tree species mapping in temperate mountain forests. bioRxiv. https://doi.org/10.64898/2026.02.23.707022

Becker, A., Wegner, J.D., Dawoe, E., Schindler, K., Thompson, W.J., Bunn, C., Garrett, R.D., Castro, F., Hart, S.P., Blaser-Hart, W.J., 2025. The unrealized potential of agroforestry for an emissions-intensive agricultural commodity. Nat. Sustain. 8, 994–1003. https://doi.org/10.1038/s41893-025-01608-7

Berger, K., Herold, M., Szantoi, Z., 2025. Earth observation as enabler for implementing the EU regulation on deforestation-free products. npj Climate Action 4, 68. https://doi.org/10.1038/s44168-025-00276-9

BNETD-CIGN, 2024. Note méthodologique de la carte d'occupation du sol de Côte d'Ivoire 2020 [WWW Document]. URL https://africageoportal.maps.arcgis.com/sharing/rest/content/items/76dc18767b89472eb89e8aa54e08a6c9/data (accessed 15 May 2026).

Brown, C.F., Kazmierski, M.R., Pasquarella, V.J., Rucklidge, W.J., Samsikova, M., Zhang, C., Shelhamer, E., Lahera, E., Wiles, O., Ilyushchenko, S., Gorelick, N., Zhang, L.L., Alj, S., Schechter, E., Askay, S., Guinan, O., Moore, R., Boukouvalas, A., Kohli, P., 2025. AlphaEarth Foundations: An embedding field model for accurate and efficient global mapping from sparse label data. arXiv. https://doi.org/10.48550/arXiv.2507.22291

Clough, Y., Faust, H., Tscharntke, T., 2009. Cacao boom and bust: sustainability of agroforests and opportunities for biodiversity conservation. Conserv. Lett. 2, 197–205. https://doi.org/10.1111/j.1755-263X.2009.00072.x

CLS, Terrasphere, 2024. COPERNICUS4GEOGLAM Validation Note LandCover v2 Ivory Coast. [WWW Document]. URL https://africageoportal.maps.arcgis.com/sharing/rest/content/items/0072843eefe545b1b43b0c5b2ebec41a/data (accessed 20 June 2026)

Copernicus Land Monitoring Service, 2025. Tree Cover Density 2020 (raster 10 m), pantropical, annual - version 1. https://doi.org/10.2909/59cc02d6-ddfe-4820-83cb-345205eacec5

European Union, 2023. Regulation (EU) 2023/1115 of the European Parliament and of the Council of 31 May 2023 on the making available on the Union market and the export from the Union of certain commodities and products associated with deforestation and forest degradation and repealing Regulation (EU) No 995/2010. Off. J. Eur. Union L150, 206-247.

Feng, Z., Atzberger, C., Jaffer, S., Knezevic, J., Sormunen, S., Young, R., Lisaius, M.C., Immitzer, M., Jackson, T., Ball, J., Coomes, D.A., Madhavapeddy, A., Blake, A., Keshav, S., 2025. TESSERA: Temporal Embeddings of Surface Spectra for Earth Representation and Analysis. arXiv. https://doi.org/10.48550/arXiv.2506.20380

Feng, Z., Atzberger, C., Jaffer, S., Knezevic, J., Sormunen, S., Young, R., Lisaius, M., Immitzer, M., Jackson, T., Ball, J., Coomes, D., Madhavapeddy, A., Blake, A., Keshav, S., 2026. Applications of the TESSERA Geospatial Foundation Model to Diverse Environmental Mapping Tasks. SSRN. https://doi.org/10.2139/ssrn.6142416

Forest Data Partnership, 2025. Community models 2025a. Online. [WWW Document]. URL https://github.com/google/forest-data-partnership/blob/main/models/model_2025a/README.md (accessed 20 June 2026)

Forest Data Partnership, 2026. Community models 2025b. Online. [WWW Document]. URL https://github.com/google/forest-data-partnership/blob/main/models/model_2025b/README.md (accessed 20 June 2026)

Kalischek, N., Lang, N., Renier, C., Daudt, R.C., Addoah, T., Thompson, W., Blaser-Hart, W.J., Garrett, R., Schindler, K., Wegner, J.D., 2023. Cocoa plantations are associated with deforestation in Côte d'Ivoire and Ghana. Nat. Food 4, 384–393. https://doi.org/10.1038/s43016-023-00751-8

Lisaius, M.C., Keshav, S., Blake, A., Atzberger, C., 2026a. Embedding-based Crop Type Classification in the Groundnut Basin of Senegal. arXiv. https://doi.org/10.48550/arXiv.2601.16900

Lisaius, M.C., Blake, A., Atzberger, C., Keshav, S., 2026b. Towards improved crop type classification: A compact embedding approach suitable for small fields. ISPRS Ann. Photogramm. Remote Sens. Spatial Inf. Sci. XI-2-2026, 873–880. https://doi.org/10.5194/isprs-annals-XI-2-2026-873-2026

Liu, Z., Feng, Y., Cao, X., Peñuelas, J., Descals, A., Yang, D., Cheung, K.K., Liu, S., Lin, Z., Liu, X., Wu, X., Ma, H., Li, W., Xia, H., Jin, S., Liang, S., Liu, L., Liu, L., Su, Y., Chen, J., Wu, J., 2026. Integrating very-high-resolution imagery, Sentinel-2 time-series data, and machine learning to map shrub fractional abundance across arid and semi-arid ecosystems in China. Remote Sens. Environ. 335, 115300. https://doi.org/10.1016/j.rse.2026.115300

Lu, D., Weng, Q., 2007. A survey of image classification methods and techniques for improving classification performance. Int. J. Remote Sens. 28, 823-870. https://doi.org/10.1080/01431160600746456

Ma, Y., Shen, Y., Swatantran, A., Lobell, D.B., 2026. Harvesting AlphaEarth: Benchmarking the geospatial foundation model for agricultural downstream tasks. Int. J. Appl. Earth Obs. Geoinf. 149, 105258. https://doi.org/10.1016/j.jag.2026.105258

Masolele, R.N., Marcos, D., De Sy, V., Abu, I.O., Verbesselt, J., Reiche, J., Herold, M., 2024. Mapping the diversity of land uses following deforestation across Africa. Sci. Rep. 14, 1681. https://doi.org/10.1038/s41598-024-52138-9

McGarigal, K., Marks, B.J., 1995. FRAGSTATS: Spatial Pattern Analysis Program for Quantifying Landscape Structure. General Technical Report PNW-GTR-351. U.S. Department of Agriculture, Forest Service, Pacific Northwest Research Station, Portland, OR. https://doi.org/10.2737/PNW-GTR-351

Möhring, J., Kattenborn, T., Mahecha, M.D., Cheng, Y., Beloiu Schwenke, M., Cloutier, M., Denter, M., Frey, J., Gassilloud, M., Göritz, A., Hempel, J., Horion, S., Jucker, T., Junttila, S., Khatri-Chhetri, P., Korznikov, K., Kruse, S., Laliberté, E., Maroschek, M., Neumeier, P., Pérez-Priego, O., Potts, A., Schiefer, F., Seidl, R., Vajna-Jehle, J., Zielewska-Büttner, K., Mosig, C., 2025. Global, multi-scale standing deadwood segmentation in centimeter-scale aerial images. ISPRS Open J. Photogramm. Remote Sens. 18, 100104. https://doi.org/10.1016/j.ophoto.2025.100104

Niether, W., Jacobi, J., Blaser, W.J., Andres, C., Armengot, L., 2020. Cocoa agroforestry systems versus monocultures: A multi-dimensional meta-analysis. Environ. Res. Lett. 15, 104085. https://doi.org/10.1088/1748-9326/abb053

Numbisi, F.N., Van Coillie, F.M.B., De Wulf, R., 2019. Delineation of cocoa agroforests using multiseason Sentinel-1 SAR images: A low grey level range reduces uncertainties in GLCM texture-based mapping. ISPRS Int. J. Geo-Inf. 8, 179. https://doi.org/10.3390/ijgi8040179

Olofsson, P., Foody, G.M., Herold, M., Stehman, S.V., Woodcock, C.E., Wulder, M.A., 2014. Good practices for estimating area and assessing accuracy of land change. Remote Sens. Environ. 148, 42–57. https://doi.org/10.1016/j.rse.2014.02.015

Orlowski, K., Meroni, M., Verhegghen, A., Sabo, F., Winchester, C., Rembold, F., and Schneider, M., 2026. COLD-CI: A large-scale very high-resolution label polygon dataset for cocoa and non-cocoa classification in Côte d'Ivoire. arXiv. https://doi.org/10.48550/arXiv.2606.20767

Renier, C., Addoah, T., Guye, V., Garrett, R., Van den Broeck, G., Zu Ermgassen, E.K.H.J., Meyfroidt, P., 2025. Direct and indirect deforestation for cocoa in the tropical moist forests of Ghana. Environ. Res.: Food Syst. 2, 025006. https://doi.org/10.1088/2976-601X/add01b

Ronneberger, O., Fischer, P., Brox, T., 2015. U-Net: Convolutional networks for biomedical image segmentation, in: Navab, N., Hornegger, J., Wells, W.M., Frangi, A.F. (Eds.), Medical Image Computing and Computer-Assisted Intervention – MICCAI 2015. Lecture Notes in Computer Science, vol. 9351. Springer, Cham, pp. 234–241. https://doi.org/10.1007/978-3-319-24574-4_28

Ruf, F., Siswoputranto, P.S. (Eds.), 1995. Cocoa Cycles: The Economics of Cocoa Supply. Woodhead Publishing, Cambridge.

Schneider, M., Winchester, C., Goldman, E., Shao, Y., 2023. Mapping cocoa and assessing deforestation risk for the cocoa sector in Côte d'Ivoire and Ghana. Technical Note. World Resources Institute, Washington, DC. https://doi.org/10.46830/writn.21.00011

Schroth, G., Läderach, P., Martinez-Valle, A.I., Bunn, C., Jassogne, L., 2016. Vulnerability to climate change of cocoa in West Africa: Patterns, opportunities and limits to adaptation. Science of the Total Environment 556, 231–241. https://doi.org/10.1016/j.scitotenv.2016.03.024

Simonetti, D., Pimple, U., Langner, A., Marelli, A., 2021. Pan-tropical Sentinel-2 cloud-free annual composite datasets. Data Brief 39, 107488. https://doi.org/10.1016/j.dib.2021.107488

Singh, C., Persson, U.M., 2026. Global patterns of commodity-driven deforestation and associated carbon emissions. Nat. Food 7, 138–151. https://doi.org/10.1038/s43016-026-01305-4

Squicciarini, M.P., Swinnen, J.F.M. (Eds.), 2016. The Economics of Chocolate. Oxford University Press, Oxford.

Tan, M., Le, Q.V., 2019. EfficientNet: Rethinking model scaling for convolutional neural networks, in: Chaudhuri, K., Salakhutdinov, R. (Eds.), Proceedings of the 36th International Conference on Machine Learning. Proc. Mach. Learn. Res. 97, 6105–6114.

Therias, A., Rafiee, A., Lhermitte, S., van der Lugt, P., Lindenbergh, R., 2025. Integrating radar and multi-spectral data to detect cocoa crops: a deep learning approach. Remote Sens. Appl.: Soc. Environ. 39, 101652. https://doi.org/10.1016/j.rsase.2025.101652

Tolan, J., Yang, H.I., Nosarzewski, B., Couairon, G., Vo, H. V., Brandt, J., Spore, J., Majumdar, S., Haziza, D., Vamaraju, J., Moutakanni, T., Bojanowski, P., Johns, T., White, B., Tiecke, T., Couprie, C., 2024. Very high resolution canopy height maps from RGB imagery using self-supervised vision transformer and convolutional decoder trained on aerial lidar. Remote Sens. Environ. 300, 113888. https://doi.org/10.1016/j.rse.2023.113888

Vaast, P., Somarriba, E., 2014. Trade-offs between crop intensification and ecosystem services: the role of agroforestry in cocoa cultivation. Agrofor. Syst. 88, 947–956. https://doi.org/10.1007/s10457-014-9762-x

Vakayil, A., Joseph, V.R., 2022. Data Twinning. Stat. Anal. Data Min. 15, 598–610. https://doi.org/10.1002/sam.11574

Zhu, Y., Dixon, D.J., Jin, Y., 2026. Scalable sub-meter mapping of woody vegetation and structures across California's heterogeneous landscape using deep learning. Remote Sens. Environ. 333, 115091. https://doi.org/10.1016/j.rse.2025.115091

# Supplementary material

**Is sub-metre resolution necessary for cocoa mapping? A landscape-stratified evaluation of very high resolution imagery, decametric Earth Observation inputs, and operational products in Côte d'Ivoire**

Kasimir Orlowski[1,2], Filip Sabo[3,*], Michele Meroni[4], Astrid Verhegghen[5,6], Mariana Belgiu[1], Felix Rembold[3]

[1]Faculty of Geo-Information Science and Earth Observation (ITC), University of Twente, Hallenweg 8, 7522 NH Enschede, The Netherlands.
[2]FINCONS S.P.A., Via Torri Bianche 10, Palazzo Betulla, 20871 Vimercate, Italy.
[3]European Commission, Joint Research Centre (JRC), Via Enrico Fermi, 2749, 21027 Ispra, Italy.
[4]Seidor Consulting, Carrer de Veneçuela 74, 08019 Barcelona, Spain.
[5]ARHS Developments Italia S.R.L., Via Fratelli Gabba 1/A, 20121 Milan, Italy.
[6]Flemish Institute for Technological Research (VITO), Boeretang 200, 2400 Mol, Belgium.

*Corresponding author: filip.szabo@ec.europa.eu

## S1. Canopy Height Threshold for Overstory Tree Identification

To identify cocoa areas with overstory tree cover, a canopy height threshold of 8 m was adopted. The threshold was informed by four complementary sources of evidence.

First, 10,000 circular polygons with a diameter of 30 m were initially placed at random within reference cocoa polygons representing monoculture cocoa systems. Each polygon was visually inspected using very high resolution (VHR) imagery to confirm the absence of shade trees. Polygons containing a mixture of cocoa and overstory trees were manually relocated to nearby monoculture cocoa areas where possible; otherwise, they were discarded. This procedure resulted in 6,992 retained polygons. Canopy height values were extracted for all canopy height model (CHM) pixels located within the 6,992 retained polygons. The 99th percentile of the pooled canopy-height distribution was below 8 m.

Second, an independent study by Becker et al. (2025) used drone-derived canopy height data collected across 827 cocoa farms distributed over four agroclimatic zones in Côte d'Ivoire and Ghana. Their analysis similarly found that 99.7% of cocoa tree height measurements were below 8 m, providing independent support for the threshold.

Third, field measurements collected within the Cocoa4Future project (Cocoa4Future, 2022) provide an independent benchmark for non-cocoa overstory vegetation. The dataset comprises 12,326 shade trees across 150 cocoa agroforestry farms and reports a mean shade tree height of 10.76 m, substantially exceeding typical cocoa tree height.

Finally, the threshold is consistent with field observations from the Copernicus4GEOGLAM validation dataset for Côte d'Ivoire (CLS and Terrasphere, 2024). Across 595 cocoa reference locations, the average reported cocoa tree height was 3.8 m, with a maximum observed height of 8 m.

Together, these complementary sources indicate that canopy heights exceeding 8 m are unlikely to represent cocoa trees and can therefore be used as a proxy for overstory tree cover within cocoa production systems in Côte d'Ivoire.

## S2. Stratified Sampling Design and Precision Assessment

**Table S1.** Per-stratum sample sizes, area-based weights, and maximum confidence-interval half-widths for the stratified accuracy assessment. The "maximum CI half-width" column reports the largest Wilson confidence-interval half-width among all evaluated models and products within each stratum. Strata where the ±5% precision criterion was not met correspond to strata where FDaP v2025b was added after reference data collection had been completed (see Section 2.6). TCD = Tree Cover Density; PD = Patch Density. CI = 95% Wilson confidence-interval half-width.

| Stratum | TCD class | PD class | n | Weight | Model/product determining maximum half-width | Maximum CI half-width |
|---|---|---|---|---|---|---|
| 11 | <30% | Low | 189 | 0.1977 | FDaP v2025b | 0.061 |
| 12 | <30% | Medium | 268 | 0.1515 | FDaP v2025b | 0.053 |
| 13 | <30% | High | 336 | 0.1061 | FDaP v2025b | 0.051 |
| 21 | 30–70% | Low | 294 | 0.0215 | BNETD | 0.050 |
| 22 | 30–70% | Medium | 357 | 0.0315 | BNETD | 0.049 |
| 23 | 30–70% | High | 368 | 0.0337 | FDaP v2025b | 0.050 |
| 31 | >70% | Low | 290 | 0.1109 | FDaP v2025a | 0.050 |
| 32 | >70% | Medium | 345 | 0.1470 | FDaP v2025b | 0.050 |
| 33 | >70% | High | 374 | 0.2002 | FDaP v2025b | 0.050 |

## S3. Stratified Accuracy Assessment Results

**Table S2.** Per-stratum accuracy metrics for all evaluated models and existing cocoa mapping products. F1 = F1-score; OA = overall accuracy; UA = user's accuracy; PA = producer's accuracy. The reported 95% Wilson confidence interval and its half-width refer to OA. Strata are defined by Tree Cover Density (TCD) and Patch Density (PD), as described in Section 2.4.1.

| Stratum | TCD | PD | Source | n | F1 | OA | UA | PA | OA CI95 lower | OA CI95 upper | OA CI95 half-width |
|---|---|---|---|---|---|---|---|---|---|---|---|
| 11 | <30% | Low | VHR | 189 | 0.923 | 0.974 | 0.857 | 1.000 | 0.940 | 0.989 | 0.025 |
| 11 | <30% | Low | TESSERA | 189 | 0.839 | 0.947 | 0.812 | 0.867 | 0.905 | 0.971 | 0.033 |
| 11 | <30% | Low | AEF | 189 | 0.774 | 0.926 | 0.750 | 0.800 | 0.880 | 0.955 | 0.038 |
| 11 | <30% | Low | Kalischek | 189 | 0.694 | 0.921 | 0.895 | 0.567 | 0.873 | 0.951 | 0.039 |
| 11 | <30% | Low | Sentinel-2 | 189 | 0.737 | 0.921 | 0.778 | 0.700 | 0.873 | 0.951 | 0.039 |
| 11 | <30% | Low | FDaP v2025a | 189 | 0.703 | 0.884 | 0.591 | 0.867 | 0.830 | 0.922 | 0.046 |
| 11 | <30% | Low | FDaP v2025b | 189 | 0.540 | 0.757 | 0.386 | 0.900 | 0.691 | 0.812 | 0.061 |
| 11 | <30% | Low | BNETD | 189 | 0.596 | 0.878 | 0.630 | 0.567 | 0.824 | 0.918 | 0.047 |
| 12 | <30% | Medium | VHR | 268 | 0.901 | 0.940 | 0.820 | 1.000 | 0.905 | 0.963 | 0.029 |
| 12 | <30% | Medium | TESSERA | 268 | 0.848 | 0.907 | 0.761 | 0.959 | 0.866 | 0.936 | 0.035 |
| 12 | <30% | Medium | AEF | 268 | 0.826 | 0.892 | 0.734 | 0.945 | 0.849 | 0.924 | 0.037 |
| 12 | <30% | Medium | Kalischek | 268 | 0.806 | 0.907 | 0.929 | 0.712 | 0.866 | 0.936 | 0.035 |
| 12 | <30% | Medium | Sentinel-2 | 268 | 0.761 | 0.862 | 0.720 | 0.808 | 0.816 | 0.898 | 0.041 |
| 12 | <30% | Medium | FDaP v2025a | 268 | 0.681 | 0.780 | 0.562 | 0.863 | 0.726 | 0.825 | 0.049 |
| 12 | <30% | Medium | FDaP v2025b | 268 | 0.657 | 0.728 | 0.500 | 0.959 | 0.671 | 0.777 | 0.053 |
| 12 | <30% | Medium | BNETD | 268 | 0.561 | 0.784 | 0.627 | 0.507 | 0.730 | 0.829 | 0.049 |
| 13 | <30% | High | VHR | 336 | 0.914 | 0.952 | 0.842 | 1.000 | 0.924 | 0.970 | 0.023 |
| 13 | <30% | High | TESSERA | 336 | 0.832 | 0.902 | 0.732 | 0.965 | 0.865 | 0.929 | 0.032 |
| 13 | <30% | High | AEF | 336 | 0.714 | 0.818 | 0.594 | 0.894 | 0.774 | 0.856 | 0.041 |
| 13 | <30% | High | Kalischek | 336 | 0.750 | 0.893 | 0.915 | 0.635 | 0.855 | 0.922 | 0.033 |
| 13 | <30% | High | Sentinel-2 | 336 | 0.682 | 0.839 | 0.682 | 0.682 | 0.796 | 0.875 | 0.039 |
| 13 | <30% | High | FDaP v2025a | 336 | 0.581 | 0.679 | 0.434 | 0.882 | 0.627 | 0.726 | 0.050 |
| 13 | <30% | High | FDaP v2025b | 336 | 0.566 | 0.631 | 0.403 | 0.953 | 0.578 | 0.681 | 0.051 |
| 13 | <30% | High | BNETD | 336 | 0.376 | 0.723 | 0.438 | 0.329 | 0.673 | 0.768 | 0.048 |
| 21 | 30–70% | Low | VHR | 294 | 0.958 | 0.935 | 0.919 | 1.000 | 0.901 | 0.958 | 0.028 |
| 21 | 30–70% | Low | TESSERA | 294 | 0.928 | 0.891 | 0.900 | 0.958 | 0.850 | 0.922 | 0.036 |
| 21 | 30–70% | Low | AEF | 294 | 0.925 | 0.884 | 0.882 | 0.972 | 0.843 | 0.916 | 0.037 |

| 21 | 30–70% | Low | Kalischek | 294 | 0.909 | 0.871 | 0.936 | 0.884 | 0.828 | 0.904 | 0.038 |
|---|---|---|---|---|---|---|---|---|---|---|---|
| 21 | 30–70% | Low | Sentinel-2 | 294 | 0.866 | 0.803 | 0.858 | 0.874 | 0.753 | 0.844 | 0.045 |
| 21 | 30–70% | Low | FDaP v2025a | 294 | 0.864 | 0.789 | 0.817 | 0.916 | 0.739 | 0.832 | 0.046 |
| 21 | 30–70% | Low | FDaP v2025b | 294 | 0.880 | 0.806 | 0.804 | 0.972 | 0.757 | 0.847 | 0.045 |
| 21 | 30–70% | Low | BNETD | 294 | 0.815 | 0.741 | 0.856 | 0.777 | 0.689 | 0.788 | 0.050 |
| 22 | 30–70% | Medium | VHR | 357 | 0.928 | 0.902 | 0.873 | 0.991 | 0.867 | 0.929 | 0.031 |
| 22 | 30–70% | Medium | TESSERA | 357 | 0.901 | 0.866 | 0.849 | 0.961 | 0.826 | 0.897 | 0.035 |
| 22 | 30–70% | Medium | AEF | 357 | 0.880 | 0.832 | 0.811 | 0.961 | 0.790 | 0.867 | 0.039 |
| 22 | 30–70% | Medium | Kalischek | 357 | 0.883 | 0.854 | 0.904 | 0.864 | 0.814 | 0.887 | 0.037 |
| 22 | 30–70% | Medium | Sentinel-2 | 357 | 0.858 | 0.812 | 0.831 | 0.886 | 0.769 | 0.849 | 0.040 |
| 22 | 30–70% | Medium | FDaP v2025a | 357 | 0.827 | 0.751 | 0.742 | 0.934 | 0.703 | 0.793 | 0.045 |
| 22 | 30–70% | Medium | FDaP v2025b | 357 | 0.815 | 0.723 | 0.710 | 0.956 | 0.674 | 0.767 | 0.046 |
| 22 | 30–70% | Medium | BNETD | 357 | 0.685 | 0.650 | 0.805 | 0.596 | 0.599 | 0.698 | 0.049 |
| 23 | 30–70% | High | VHR | 368 | 0.913 | 0.910 | 0.845 | 0.994 | 0.877 | 0.935 | 0.029 |
| 23 | 30–70% | High | TESSERA | 368 | 0.869 | 0.864 | 0.802 | 0.949 | 0.825 | 0.895 | 0.035 |
| 23 | 30–70% | High | AEF | 368 | 0.822 | 0.810 | 0.740 | 0.926 | 0.767 | 0.847 | 0.040 |
| 23 | 30–70% | High | Kalischek | 368 | 0.835 | 0.845 | 0.847 | 0.823 | 0.805 | 0.878 | 0.037 |
| 23 | 30–70% | High | Sentinel-2 | 368 | 0.745 | 0.739 | 0.697 | 0.800 | 0.692 | 0.781 | 0.045 |
| 23 | 30–70% | High | FDaP v2025a | 368 | 0.693 | 0.620 | 0.562 | 0.903 | 0.569 | 0.668 | 0.049 |
| 23 | 30–70% | High | FDaP v2025b | 368 | 0.680 | 0.573 | 0.528 | 0.954 | 0.522 | 0.623 | 0.050 |
| 23 | 30–70% | High | BNETD | 368 | 0.495 | 0.606 | 0.634 | 0.406 | 0.555 | 0.655 | 0.050 |
| 31 | >70% | Low | VHR | 290 | 0.944 | 0.962 | 0.894 | 1.000 | 0.933 | 0.979 | 0.023 |
| 31 | >70% | Low | TESSERA | 290 | 0.900 | 0.931 | 0.841 | 0.968 | 0.896 | 0.955 | 0.030 |
| 31 | >70% | Low | AEF | 290 | 0.879 | 0.914 | 0.798 | 0.978 | 0.876 | 0.941 | 0.033 |
| 31 | >70% | Low | Kalischek | 290 | 0.888 | 0.931 | 0.929 | 0.849 | 0.896 | 0.955 | 0.030 |
| 31 | >70% | Low | Sentinel-2 | 290 | 0.778 | 0.862 | 0.805 | 0.753 | 0.818 | 0.897 | 0.040 |
| 31 | >70% | Low | FDaP v2025a | 290 | 0.689 | 0.748 | 0.570 | 0.871 | 0.695 | 0.795 | 0.050 |
| 31 | >70% | Low | FDaP v2025b | 290 | 0.706 | 0.759 | 0.579 | 0.903 | 0.706 | 0.804 | 0.049 |

| 31 | >70% | Low | BNETD | 290 | 0.701 | 0.817 | 0.738 | 0.667 | 0.769 | 0.857 | 0.044 |
|---|---|---|---|---|---|---|---|---|---|---|---|
| 32 | >70% | Medium | VHR | 345 | 0.907 | 0.919 | 0.835 | 0.993 | 0.885 | 0.943 | 0.029 |
| 32 | >70% | Medium | TESSERA | 345 | 0.838 | 0.861 | 0.785 | 0.899 | 0.820 | 0.893 | 0.037 |
| 32 | >70% | Medium | AEF | 345 | 0.814 | 0.832 | 0.730 | 0.920 | 0.789 | 0.868 | 0.039 |
| 32 | >70% | Medium | Kalischek | 345 | 0.835 | 0.872 | 0.867 | 0.804 | 0.833 | 0.904 | 0.035 |
| 32 | >70% | Medium | Sentinel-2 | 345 | 0.758 | 0.812 | 0.779 | 0.739 | 0.767 | 0.849 | 0.041 |
| 32 | >70% | Medium | FDaP v2025a | 345 | 0.655 | 0.655 | 0.546 | 0.819 | 0.603 | 0.703 | 0.050 |
| 32 | >70% | Medium | FDaP v2025b | 345 | 0.674 | 0.641 | 0.529 | 0.928 | 0.589 | 0.689 | 0.050 |
| 32 | >70% | Medium | BNETD | 345 | 0.564 | 0.672 | 0.603 | 0.529 | 0.621 | 0.720 | 0.049 |
| 33 | >70% | High | VHR | 374 | 0.910 | 0.928 | 0.834 | 1.000 | 0.897 | 0.950 | 0.027 |
| 33 | >70% | High | TESSERA | 374 | 0.841 | 0.872 | 0.765 | 0.934 | 0.834 | 0.902 | 0.034 |
| 33 | >70% | High | AEF | 374 | 0.822 | 0.856 | 0.744 | 0.919 | 0.816 | 0.888 | 0.036 |
| 33 | >70% | High | Kalischek | 374 | 0.838 | 0.885 | 0.860 | 0.816 | 0.849 | 0.914 | 0.032 |
| 33 | >70% | High | Sentinel-2 | 374 | 0.720 | 0.802 | 0.742 | 0.699 | 0.759 | 0.839 | 0.040 |
| 33 | >70% | High | FDaP v2025a | 374 | 0.567 | 0.575 | 0.450 | 0.765 | 0.524 | 0.624 | 0.050 |
| 33 | >70% | High | FDaP v2025b | 374 | 0.567 | 0.551 | 0.437 | 0.809 | 0.500 | 0.600 | 0.050 |
| 33 | >70% | High | BNETD | 374 | 0.402 | 0.650 | 0.530 | 0.324 | 0.600 | 0.696 | 0.048 |

## S4. Spectral Ablation Experiment

**Table S3.** Area-adjusted accuracy metrics derived from the stratified random sample of reference points. F1 = F1-score; OA = overall accuracy; UA = user's accuracy; PA = producer's accuracy. The OA CI95 half-width is the half-width of the 95% confidence interval for area-adjusted OA, estimated following Olofsson et al. (2014). S2_4b denotes the Sentinel-2 model trained using only the blue, green, red, and near-infrared bands.

| EO input | F1 | OA | UA | PA | OA CI95 half-width |
|---|---|---|---|---|---|
| VHR | 0.92 | 0.94 | 0.85 | 1.00 | 0.010 |
| Sentinel-2 | 0.76 | 0.85 | 0.76 | 0.75 | 0.016 |
| S2_4b | 0.64 | 0.76 | 0.62 | 0.66 | 0.018 |

## References

Becker, A., Wegner, J.D., Dawoe, E., Schindler, K., Thompson, W.J., Bunn, C., Garrett, R.D., Castro, F., Hart, S.P., Blaser-Hart, W.J., 2025. The unrealized potential of agroforestry for an emissions-intensive agricultural commodity. Nat. Sustain. 8, 994–1003. https://doi.org/10.1038/s41893-025-01608-7

CLS, Terrasphere, 2024. COPERNICUS4GEOGLAM Validation Note LandCover v2 Ivory Coast. [WWW Document]. URL https://africageoportal.maps.arcgis.com/sharing/rest/content/items/0072843eefe545b1b43b0c5b2ebec41a/data (accessed 20 June 2026)

Cocoa4Future, 2022. Cocoa4Future project database of shade-tree measurements from cocoa agroforestry systems in Côte d'Ivoire. CIRAD, Montpellier, France. Unpublished dataset.

Olofsson, P., Foody, G.M., Herold, M., Stehman, S.V., Woodcock, C.E., Wulder, M.A., 2014. Good practices for estimating area and assessing accuracy of land change. Remote Sens. Environ. 148, 42–57. https://doi.org/10.1016/j.rse.2014.02.015